# Prediction of Malignant & Benign Breast Cancer: A Data Mining Approach in Healthcare Applications

Vivek Kumar[1] [0000-0003-3958-4704], Brojo Kishore Mishra[2] [0000-0002-7836-052X], Manuel Mazzara[3] [0000-0002-3860-4948], Dang N. H. Thanh[4] [0000-0003-2025-8319], Abhishek Verma[5]

[1] National University of Science and Technology-MiSiS, Moscow, Russian Federation, `vivekumar0416@gmail.com`
[2] GIET University, Gunupur, India, `brojomishra@gmail.com`
[3] Innopolis University, Kazan, Russian Federation, `m.mazzara@innopolis.ru`
[4] Hue College of Industry, Vietnam, `dnhthanh@hueic.edu.vn`
[5] Malaviya National Institute of Technology, Jaipur, India, `iamvermaabhishek@gmail.com`

**Abstract.** As much as data science is playing a pivotal role everywhere, healthcare also finds it prominent application. Breast Cancer is the top rated type of cancer amongst women; which took away 627,000 lives alone. This high mortality rate due to breast cancer does need attention, for early detection so that prevention can be done in time. As a potential contributor to state-of-art technology development, data mining finds a multi-fold application in predicting Brest cancer. This work focuses on different classification techniques implementation for data mining in predicting malignant and benign breast cancer. Breast Cancer Wisconsin data set from the UCI repository has been used as experimental dataset while attribute clump thickness being used as evaluation class. The performances of these twelve algorithms: Ada Boost M1, Decision Table, J-Rip, J48, Lazy IBK, Lazy K-star, Logistics Regression, Multiclass Classifier, Multilayer–Perceptron, Naïve Bayes, Random Forest and Random Tree is analyzed on this data set.

**Keywords:** Data Mining, Classification Techniques, UCI repository, Breast Cancer, Classification Algorithms

## 1 Introduction

"People used to say everyone knows someone who's had breast cancer. In the past few weeks, I've learned something else: Everyone has someone close to them who has had breast cancer.-Debbie Wasserman Schultz, US House of Representatives, breast cancer survivor. Of the 184 major countries in the world, breast cancer is the most common cancer diagnosis in women in 140 countries (76%) and the most frequent cause of cancer mortality in 101 countries (55%) [1]. Breast cancer ratio are statistically higher in women in more developed countries as compared to other diseases. But it is also globally increasing day by day.

Table 1 Shows the ranking of top twenty-five countries most affected by breast cancer [2]. To discourage the growth of breast cancer, it is important to focus on early detection. Early diagnosis and screening are two main methods of advance detection

of breast cancer. The poor regions can be made aware by familiarizing with early diagnosis program, as state by World Health Organization. It includes early diagnosis, screening, mammography and Clinical Breast Exam (CBE) [3].

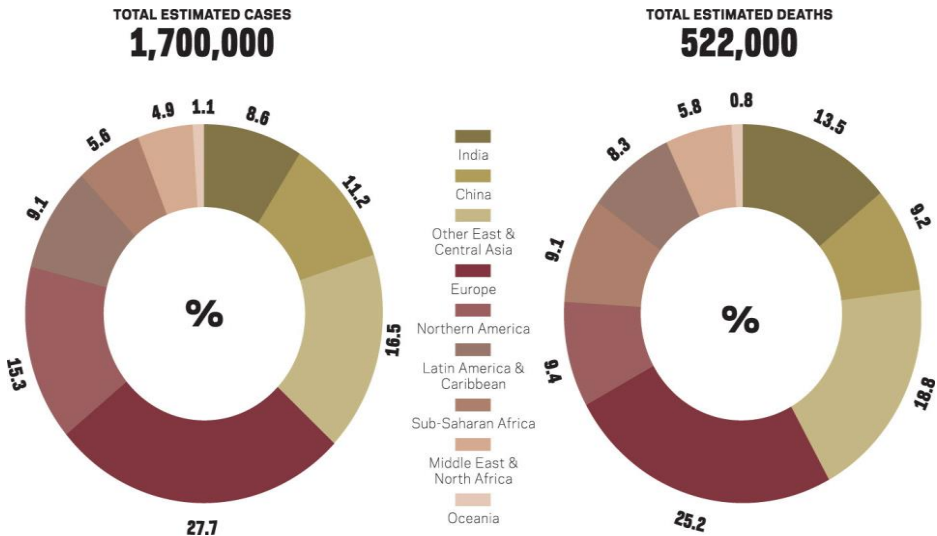

**Fig. 1.** Estimated new breast cancer cases and deaths by region.

**Table 1.** Estimated new breast cancer cases and deaths by region.

| Rank | Country | Age-Standardized Rate Per 100,000 |
|------|---------|-----------------------------------|
| 1 | Belgium | 113.2 |
| 2 | Luxembourg | 109.3 |
| 3 | Netherlands | 105.9 |
| 4 | France (metropolitan) | 99.1 |
| 5 | New Caledoni (France) | 98.0 |
| 6 | Lebanon | 97.6 |
| 7 | Australia | 94.5 |
| 8 | UK | 93.6 |
| 9 | Italy | 92.8 |
| 10 | New Zealand | 92.6 |
| 11 | Ireland | 90.3 |
| 12 | Sweden | 89.8 |
| 13 | Finland | 89.5 |
| 14 | Denmark | 88.8 |
| 15 | Switzerland | 88.1 |
| 16 | Montenegro | 87.8 |
| 17 | Malta | 87.6 |
| 18 | Norway | 87.5 |
| 19 | Hungary | 85.5 |
| 20 | Germany | 85.4 |
| 21 | Iceland | 85.2 |
| 22 | US | 84.9 |
| 23 | Canada | 83.8 |
| **24** | Cyprus | 81.7 |
| **25** | Samoa | 80.1 |

## 2    Related Work

Data have become the new center of power. The quantum leap in data analysis application in all fields has empowered the use of data mining. Data mining find its way in finding trends and deviations of patients, , spotting trends in Twitter, road accidents, face detection, counter-terrorism, pattern analysis in bioinformatics agriculture, fraud detection etc. [4-8].

Chaurasia V, discusses the use of available technological advancements to develop prediction models for breast cancer. The manuscript used Naïve Bayes, RBF Network and J48 to develop prediction model by 10 fold cross-validation method for measuring the unbiased estimate of these models for performance comparison [9]. Verma D, used five classification algorithms Naïve bayes, SMO, REP Tree, J48 and MLP upon two data sets which are breast cancer and diabetes respectively, from the UCI machine learning repository [10].

Ojha. U, emphasize upon the selection parameters for predicting the probability of recurrence of breast cancer by using data mining techniques. The manuscripts depict the deployment of clustering and classification algorithms and. The author says that classification algorithms worked better than clustering for the experimental data set. The clustering algorithms used were K-Means, EM, PAM, Fuzzy c-mean and Mean while classification algorithms were C 5.0, KNN, Naïve Bayes, SVM and Mean [11].

Rodrigues., B.L, compares two machine learning techniques to create classifier that can discriminate benign from malignant breast lumps. Wisconsin Breast Cancer Diagnosis data set is used for this purpose. The manuscript also discusses the insight of data and how to deal with the missing values and avoid overfitting or underfitting of the implemented classifiers [12]. Saxena S, implemented neural network for classification of breast cancer data. The paper studies various techniques used for the diagnosis of breast cancer using ANN and discusses its accuracy [13].

## 3    Methodology

Twelve classification algorithms were performed in this work namely; Ada Boost M1, Decision Table, J-Rip, J48, Lazy IBK, Lazy K-star, Logistics Regression, Multiclass Classifier, Multilayer–Perceptron, Naïve Bayes, Random Forest and Random Tree. The attribute Clump Thickness was take as evaluation class. Ten attributes of the data set were exercised while experiment. Table 2 depicts the attributes of the data set. For attribute class- numeric value 2 and 4 were set to signify benign and malignant tumors. Wisconsin Breast Cancer Diagnosis dataset from UCI repository and other public domain available data set are used to train the model [13-18]. A brief notes about the parameters is presented below to enumerate the results findings of the implemented classification algorithms.

Table 2. Attributes taken for data set.

| Sr. No | Attributes | Scales |
|---|---|---|
| 1 | Sample code number | ID number |
| 2 | Clump Thickness | 1 - 10 |
| 3 | Uniformity of Cell Size | 1 - 10 |
| 4 | Uniformity of Cell Shape | 1 - 10 |
| 5 | Marginal Adhesion | 1 - 10 |
| 6 | Single Epithelial Cell Size | 1 - 10 |
| 7 | Bare Nuclei | 1 - 10 |
| 8 | Bland Chromatin | 1 - 10 |
| 9 | Normal Nucleoli | 1 - 10 |
| 10 | Mitoses | 1 - 10 |
| 11 | Class | 2 (Benign) , 4 (Malignant) |

- Kappa Statistics- The Kappa statistic (or value) is a measure which makes a comparison between expected accuracy and observed accuracy. It is used not only to evaluate a single classifier, but also to evaluate classifiers amongst themselves. A value greater than 0 means classifier is doing better, while 0 represents the result is same as expected.

$$k = (p_o - p_e)/(1 - p_e) = 1 - (1 - p_o/1 - p_e) \qquad (1)$$

- Mean absolute error- It is a measure of difference between two continuous variables. The mean absolute error is given by the equation:

$$MAE = 1/n \sum_{i=1}^{n} |f_i - y_i| = 1/n \sum_{i=1}^{n} |e_i| \qquad (2)$$

- Root mean squared error-It is a metric of the differences between the values predicted by a model and the actually observed values. It represents the sample standard deviation of the differences between predicted values and observed values. The RMSD of predicted values for (y_t )ˇtimes t of a regression's dependent variable y is computed for n different predictions as the square root of the mean of the squares of the deviations:

$$RMSD = \sqrt{\sum_{t=1}^{n} (\widetilde{y_t} - y)^2/n} \qquad (3)$$

- Relative absolute error- The relative error is the absolute error divided by the magnitude of the exact value. The percent error is the relative error expressed in terms of per 10. The equation of RAE is as given below:

$$MAE = \sum_{i=1}^{n} |p_i - a_i| / \sum_{i=1}^{n} |\bar{a} - a_i| \qquad (4)$$

## 4 Results and Discussion

Figure 2 shows the results of the classifiers implemented. The classification accuracy percentage is achieved in three categories which a) 71-80 %, b) 81-90% and c) 90-100%. Only Naïve Bayes is in category a, while J48, Ada Boost-M1, Decision and J-Rip are in category b. From Multiclass to Lazy IBK have performed exceptionally well and falls in category c. Table 3 sums up the Kappa Statistics, MAE, RMS, RAE, RRSE values of the implemented classifiers in this work. Table 4 shows the PRC Area, ROC Area, MCC, F-Measure, Recall, Precision, FP Rate, TP Rate value of all the 12 classifiers.

**Table 3.** Kappa Statistics, MAE, RMS, RAE, RRSE values of the implemented classifiers.

| Classifier | Kappa Statistics | MAE | RMSE | RAE | RRSE |
|---|---|---|---|---|---|
| Ada Boost M1 | 0 | 0.1482 | 0.2425 | 211.21% | 130.95% |
| Decision Table | 0.0594 | 0.0705 | 0.1736 | 100.47% | 93.78% |
| J-Rip | 0.3306 | 0.0558 | 0.1671 | 79.55% | 90.24% |
| J48 | 0 | 0.0685 | 0.1851 | 97.66% | 99.98% |
| Lazy IBK | 0.9715 | 0.0049 | 0.0406 | 7.01% | 21.95% |
| Lazy K-star | 0.9715 | 0.0042 | 0.0412 | 6.02% | 22.24% |
| Logistics Regression | 0.9029 | 0.0128 | 0.0763 | 18.21% | 41.20% |
| Multiclass Classifier | 0.7973 | 0.025 | 0.1054 | 35.55% | 56.92% |
| Multilayer -Perceptron | 0.9142 | 0.0073 | 0.0687 | 10.46% | 37.08% |
| Naïve Bayes | 0.402 | 0.0627 | 0.2004 | 89.32% | 108.22% |
| Random Forest | 0.9715 | 0.0233 | 0.0748 | 33.15% | 40.39% |
| Random Tree | 0.9715 | 0.0033 | 0.0405 | 4.68% | 21.88% |

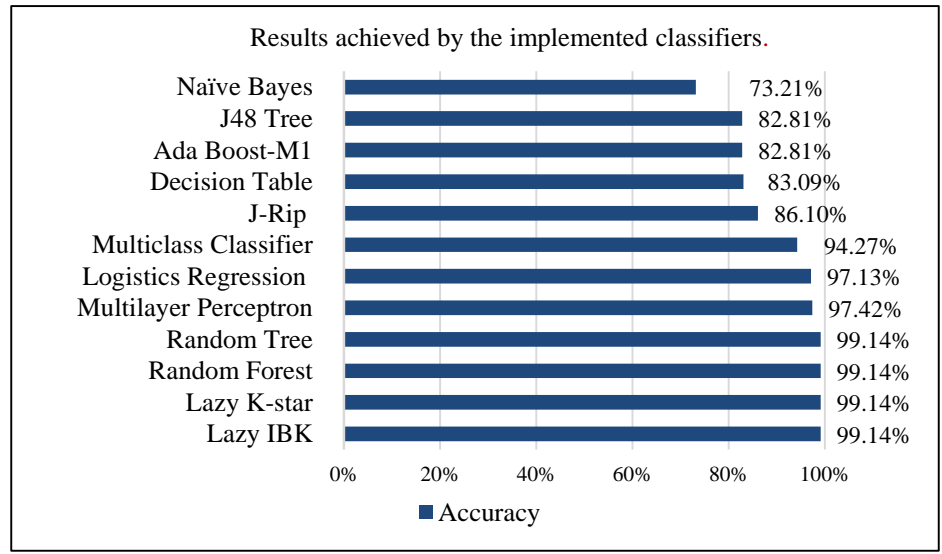

**Fig. 2.** Results achieved by the implemented classifiers

Table 4. Result of Classifiers

| Classifier | TP Rate | FP Rate | Precision | Recall | F-Measure | MCC | ROC Area | PRC Area |
|---|---|---|---|---|---|---|---|---|
| Ada Boost M1 | 0.828 | 0.828 | 0.686 | 0.828 | 0.750 | 0.000 | 0.823 | 0.791 |
| Decision Table | 0.831 | 0.787 | 0.713 | 0.831 | 0.760 | 0.160 | 0.851 | 0.819 |
| J-Rip | 0.861 | 0.635 | 0.811 | 0.861 | 0.814 | 0.402 | 0.617 | 0.754 |
| J48 | 0.828 | 0.828 | 0.686 | 0.828 | 0.750 | 0.000 | 0.500 | 0.692 |
| Lazy IBK | 0.991 | 0.041 | 0.991 | 0.991 | 0.991 | 0.970 | 0.998 | 0.996 |
| Lazy K-star | 0.991 | 0.041 | 0.991 | 0.991 | 0.991 | 0.970 | 0.998 | 0.996 |
| Logistics Regression | 0.971 | 0.111 | 0.970 | 0.971 | 0.970 | 0.899 | 0.985 | 0.985 |
| Multiclass Classifier | 0.943 | 0.215 | 0.939 | 0.943 | 0.938 | 0.795 | 0.973 | 0.970 |
| Multilayer Perceptron | 0.974 | 0.076 | 0.974 | 0.974 | 0.974 | 0.921 | 0.950 | 0.965 |
| Naïve Bayes | 0.732 | 0.100 | 0.854 | 0.732 | 0.771 | 0.493 | 0.897 | 0.880 |
| Random Forest | 0.991 | 0.041 | 0.991 | 0.991 | 0.991 | 0.970 | 0.989 | 0.992 |
| Random Tree | 0.991 | 0.041 | 0.991 | 0.991 | 0.991 | 0.970 | 0.998 | 0.996 |

## 5 Conclusion and Future Work

Using prediction model to classify cases of breast cancer is statistical in nature. In this paper twelve distinct machine learning techniques for breast cancer diagnosis were used. The results are promising and most of them scores above 94%. Only Naïve Bayes has underperformed compared to other with accuracy of 73.21%. Tree and Lazy classifier algorithms have performed exceptionally well; accuracy being close to 99%.

As extension of this work; future work includes the implementation of artificial neural net and deep learning for predictive model development with a larger and unstructured data set. This will use unsupervised learning algorithms such dimensionality reduction PCA, SVM etc. to first label the data and distributing them over training set, cross-validation set and test set.

## Abbreviations

Mean absolute error-MAE, Root mean squared error-RMSE, Relative absolute error-RAE, Root relative squared error- RRSE, TP- True Positive, TN- True Negative.